\documentclass[letterpaper, 10 pt, conference]{ieeeconf}  %

\IEEEoverridecommandlockouts                              %

\usepackage{graphicx}
\usepackage{xcolor,balance}
\usepackage{amsmath}
\usepackage{amssymb}

\usepackage{silence}
\usepackage[font=footnotesize]{caption}
\usepackage{subcaption}

\usepackage{bm} %

\let\labelindent\relax
\usepackage{enumitem}
\setlist{nosep}
\setlist{noitemsep}
\setlist{nolistsep}

\usepackage{multirow}
\usepackage{multicol}
\usepackage{colortbl}

\usepackage{algorithm}
\usepackage{algpseudocode}

\usepackage{booktabs}

\usepackage[pdftex, pdfstartview={FitV}, pdfpagelayout={TwoColumnLeft},bookmarksopen=true,plainpages = false, colorlinks=true, linkcolor=black, citecolor = black, urlcolor = black,filecolor=black , pagebackref=false,hypertexnames=false, plainpages=false, pdfpagelabels ]{hyperref}
\usepackage[sort,compress]{cite}

\usepackage{textcomp}

\definecolor{tab-blue}{rgb}{0.122, 0.467, 0.706}
\definecolor{tab-orange}{rgb}{1.000, 0.498, 0.055}
\definecolor{tab-green}{rgb}{0.173, 0.627, 0.173}
\definecolor{tab-red}{rgb}{0.839, 0.153, 0.157}
\definecolor{tab-purple}{rgb}{0.580, 0.404, 0.741}
\definecolor{tab-brown}{rgb}{0.549, 0.337, 0.294}
\definecolor{tab-pink}{rgb}{0.890, 0.467, 0.761}
\definecolor{tab-gray}{rgb}{0.498, 0.498, 0.498}
\definecolor{tab-olive}{rgb}{0.737, 0.741, 0.133}
\definecolor{tab-cyan}{rgb}{0.090, 0.745, 0.812}

\newcommand{\meas}[1]{$\text{#1}$}

\usepackage{mathtools} %

\DeclarePairedDelimiter\BigP{\Big(}{\Big)}
\DeclarePairedDelimiter\BiggP{\Bigg(}{\Bigg)}

\newcommand\scalemath[2]{\scalebox{#1}{\mbox{\ensuremath{\displaystyle #2}}}}

\newcommand{\reals}{\mathbb{R}}

\DeclarePairedDelimiter\tup{\langle}{\rangle}

\newcommand{\I}{\mathcal{I}}
\newcommand{\F}{\mathcal{F}}
\newcommand{\R}{\mathcal{R}}
\renewcommand{\P}{\mathcal{P}}
\renewcommand{\L}{\ell}

\newcommand{\SE}[1]{SE(#1)}
\newcommand{\SO}[1]{SO(#1)}

\renewcommand{\a}{A}
\renewcommand{\b}{B}
\newcommand{\w}{W}
\newcommand{\ab}{^{\a}_{\b}}

\newcommand{\A}{A}
\newcommand{\B}{B}
\newcommand{\Ai}{{\A_i}}
\newcommand{\Bj}{{\B_j}}
\newcommand{\AB}{^{\A}_{\B}}
\newcommand{\AiBj}{^\Ai_\Bj}
\newcommand{\wA}{^{\w}_{\A}}
\newcommand{\wB}{^{\w}_{\B}}
\newcommand{\wR}{^{\w}_{R}}

\newcommand{\NA}{{N_\A}}
\newcommand{\NB}{{N_\B}}

\newcommand{\p}{\mathbf{p}}
\newcommand{\ph}{\breve{\p}}

\newcommand{\T}{\mathbf{T}}
\newcommand{\Tab}{\T\ab}
\newcommand{\TAB}{\T\AB}
\newcommand{\TAiBj}{\T\AiBj}

\newcommand{\Rot}{\mathbf{R}}

\newcommand{\Rotab}{\Rot\ab}

\newcommand{\x}{x}
\newcommand{\xab}{\x\ab}
\newcommand{\xAB}{\x\AB}

\newcommand{\y}{y}
\newcommand{\yab}{\y\ab}
\newcommand{\yAB}{\y\AB}

\newcommand{\z}{z}
\newcommand{\zab}{\z\ab}
\newcommand{\zAB}{\z\AB}

\newcommand{\roll}{\alpha}
\newcommand{\rollab}{\roll\ab}
\newcommand{\rollAB}{\roll\AB}

\newcommand{\pitch}{\beta}
\newcommand{\pitchab}{\pitch\ab}
\newcommand{\pitchAB}{\pitch\AB}

\newcommand{\yaw}{\gamma}
\newcommand{\yawab}{\yaw\ab}
\newcommand{\yawAB}{\yaw\AB}

\newcommand{\trans}{\mathbf{t}}
\newcommand{\transab}{\trans\ab}

\newcommand{\envel}{\bm{\kappa}}
\newcommand{\envelc}{\hat{\envel}}
\newcommand{\envelt}{\check{\envel}}
\newcommand{\envelm}{\tilde{\envel}}

\newcommand{\zc}{\hat{\z}}
\newcommand{\rollc}{\hat{\roll}}
\newcommand{\pitchc}{\hat{\pitch}}

\newcommand{\zt}{\check{\z}}
\newcommand{\rollt}{\check{\roll}}
\newcommand{\pitcht}{\check{\pitch}}

\newcommand{\zm}{\tilde{\z}}
\newcommand{\rollm}{\tilde{\roll}}
\newcommand{\pitchm}{\tilde{\pitch}}

\renewcommand{\d}{d}
\newcommand{\dm}{\tilde{\d}}
\newcommand{\dn}{\bar{\d}}

\newcommand{\dv}{\mathbf{\d}}
\newcommand{\dvm}{\tilde{\dv}}

\newcommand{\e}{e}

\newcommand{\sumi}{\sum_{i = 1}^{\NA}}
\newcommand{\sumj}{\sum_{j = 1}^{\NB}}

\newcommand{\bs}{\vspace{1mm} \noindent}
\newcommand{\bns}{\noindent}

\newcommand{\mytexttilde}{\raisebox{0.5ex}{\texttildelow}}

\title{\LARGE \bf MURP: Multi-Agent Ultra-Wideband Relative Pose Estimation \\%
with Constrained Communications in 3D Environments}

\author{Andrew Fishberg \and Brian Quiter \and Jonathan P. How
\thanks{* Work supported in part by DOE, NNSA, and ALB funding.
A. Fishberg and J. How are with MIT Department of Aeronautics and Astronautics, \texttt{\{fishberg,jhow\}@mit.edu}.
B. Quiter is with Lawrence Berkeley National Laboratory, \texttt{bjquiter@lbl.gov}. LBNL's contribution was performed under the auspices of the US Department of Energy by Lawrence Berkeley National Laboratory under Contract DE-AC02-05CH11231. The project was funded by the US Department of Energy, National Nuclear Security Administration, Office of Defense Nuclear Nonproliferation Research and Development (DNN R\&D).}
}

\date{}
\newcommand{\comment}[1]{}

\begin{document}

\maketitle

\begin{abstract}
Inter-agent relative localization is critical for many multi-robot systems operating in the absence of external positioning infrastructure or prior environmental knowledge.
We propose a novel inter-agent relative 3D pose estimation system where each participating agent is equipped with several ultra-wideband (UWB) ranging tags.
Prior work typically supplements noisy UWB range measurements with additional
\textit{continuously} transmitted data (e.g., odometry) leading to potential scaling issues with increased team size and/or decreased communication network capability.
By equipping each agent with multiple UWB antennas, our approach addresses these concerns by using only \textit{locally} collected UWB range measurements, 
\textit{a priori} state constraints, and \textit{event-based} detections of when said constraints are violated.
The addition of our learned mean ranging bias correction improves our approach by an additional 19\% positional error, and gives us an overall experimental mean absolute position and heading errors of $\text{0.24m}$ and 9.5$\bm{^\circ}$ respectively.
When compared to other state-of-the-art approaches, our work demonstrates improved performance over similar systems, while remaining competitive with methods that have significantly higher communication costs. Additionally, we make our datasets available.

\noindent
\begin{center}
\boxed{$\footnotesize Data/Video: \url{https://github.com/mit-acl/murp-datasets}$}
\end{center}
\end{abstract}

\section{Introduction}
\label{sec:introduction}

\begin{figure}[tp]
\centering
\includegraphics[width=\linewidth]{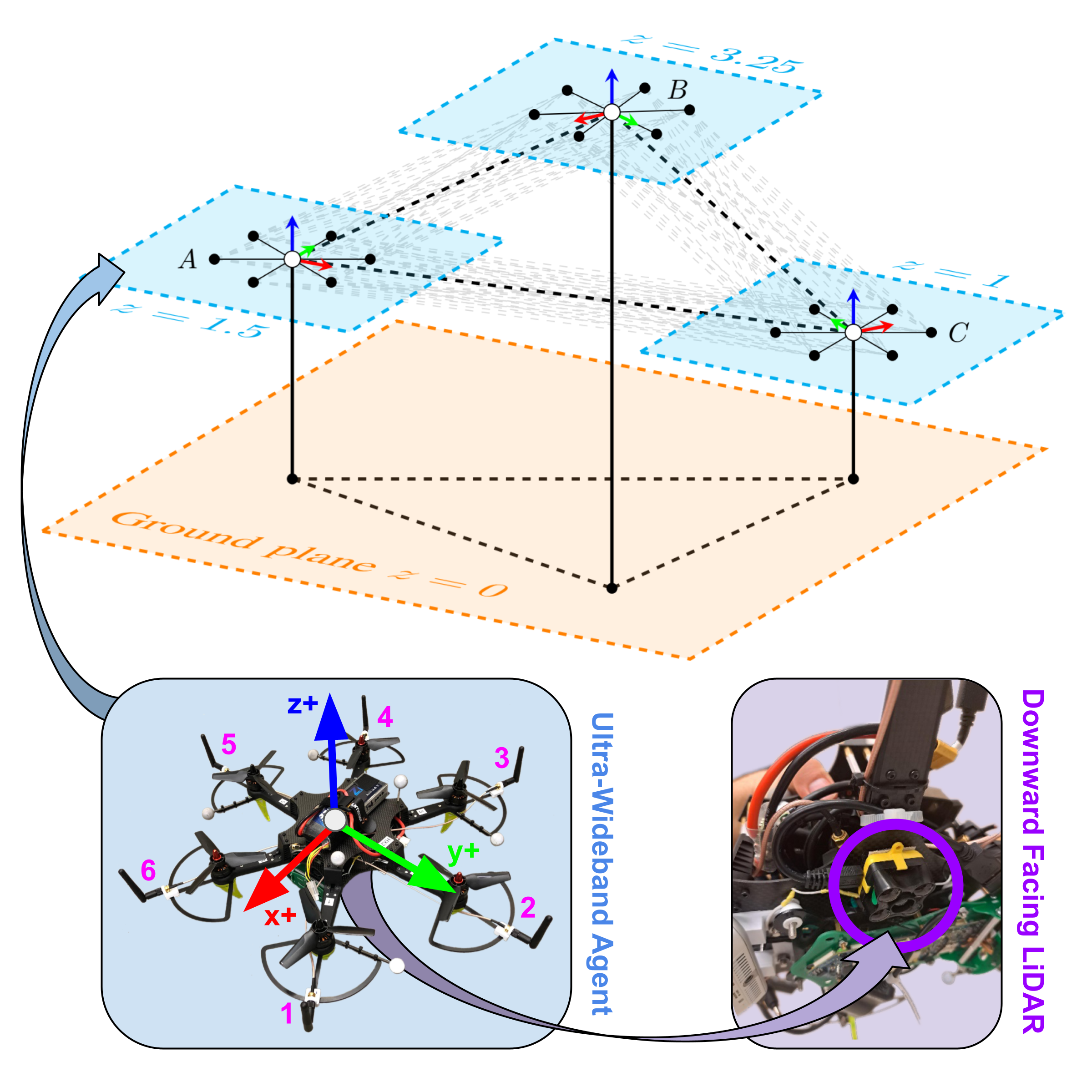}
\vspace*{-0.8cm}
\caption{Diagram of proposed system. Here three agents fly at different altitudes while performing real-time 3D relative pose estimation. Each agent is equipped with six ultra-wideband (UWB) antennas, each capable of performing pairwise relative ranging between all other agents' individual antennas. By using trilateration, an improved sensor model, and \textit{a priori} state constraints about altitude/roll/pitch, agents can perform \textit{instantaneous} estimation entirely with \textit{locally} collected UWB measurements (i.e., without the need to \textit{continuously} transmit other measurements, such as odometry). Additionally, each agent \textit{locally} monitors its \textit{a priori} constraints via downward facing LiDAR and IMU, enabling an \textit{event-based} communication model that only transmits if said assumptions change or are violated. The pictured drone is used in the experiments outlined in Section \ref{sec:uav-experiments}.}
\label{fig:system-diagram}
\vspace{-6mm}
\end{figure}

Within the last decade, ultra-wideband (UWB) has matured into a reliable, inexpensive,\footnotemark and commercially available RF solution for data transmission, relative ranging, and localization.
For robotics, UWB has several properties of note: precision of \meas{10cm}, ranges up to \meas{100m}, resilience to multipath, operates in non-line of sight (NLOS) conditions, low power consumption, and \meas{100Mbit/s} communication speeds \cite{DBLP:journals/sensors/AlarifiAAAAAA16}. Recent devices even extend the recommended and operational ranges to \meas{300m} and \meas{500m} respectively \cite{hongli_linktrack_nodate}.
Nevertheless, UWB measurements are not immune from ranging errors or noise (see Section \ref{sec:characterizing-noise}), the modeling and correction of which is an active area of research \cite{DBLP:conf/comsnets/SmaouiGK20,DBLP:conf/ccta/LedergerberD17,DBLP:journals/access/LedergerberD18,DBLP:journals/access/HamerD18}.

A common approach in UWB relative localization work fuses noisy UWB ranging measurements with additional \textit{continuously} transmitted data, such as odometry \cite{DBLP:conf/iros/Cao0YANMT21} and visual inter-agent tracks \cite{DBLP:conf/icra/XuWZQS20,DBLP:journals/corr/abs-2103-04131}.
While these approaches achieve low absolute position error (APE) and absolute heading error (AHE), there are two prevalent shortcomings:
(1)~They often use a simplistic UWB measurement noise model (i.e., zero mean Gaussian), which then requires the use of supplementary measurements to compensate.
(2)~They rely on these supplementary measurements (often not \textit{locally}\footnotemark~collected, e.g., odometry), which mandates their \textit{continuous} transmission between agents and impacts system scalability 
with increased swarm size or decreased communication throughput.

Our previous work \cite{DBLP:conf/iros/FishbergH22} used UWB to demonstrate an \textit{instantaneous}\footnotemark\ multi-tag approach to relative 2D pose estimation that achieved superior mean position accuracy and competitive performance on other metrics to Cao et al.~\cite{DBLP:conf/iros/Cao0YANMT21} (the most comparable state-of-the-art work) using only \textit{local} UWB measurements.
These results were achieved by a trilateration nonlinear least squares (NLLS) optimization problem that, leveraging an improved UWB sensor model, accounted for various sources of measurement error.

This letter extends our prior 2D work \cite{DBLP:conf/iros/FishbergH22} to an \textit{instantaneous} multi-tag relative pose estimate in 3D environments (Figure \ref{fig:system-diagram}). Similar to our previous work, a key objective is to develop an approach that minimizes the communication load (i.e., we are willing to trade absolute accuracy for reduced communication). 
In 3D environments, each agent's altitude/roll/pitch are \textit{locally} measurable with respect to a common level-ground world frame (e.g., via altimeter or downward facing LiDAR and IMU). By constraining each agent to an altitude/roll/pitch envelope, this information only needs to be transmitted once \textit{a priori}. 
These assumptions can then be \textit{locally} monitored by each agent, with \textit{event-based} 
communication only occurring when an agent changes or violates its constraint envelope (Figure \ref{fig:altitude-violation}).
This formulation allows us to simplify the 6-DoF optimization (i.e., 3D position and orientation) to a 3-DoF optimization (i.e., $x$, $y$, yaw with known $z$, roll, pitch) that, with fewer free variables, more reliably produces 3D localization solutions. Thus our approach provides a 3D relative pose using only \textit{locally} collected UWB range measurements, 
\textit{a priori} state constraints, and detections of constraint violations.

\addtocounter{footnote}{-2}
\footnotetext{At time of writing, UWB sensors sell for \mytexttilde\$10 as an integrated circuit and between \$20-\$100 as a development board or plug-and-play sensor.}

\addtocounter{footnote}{1}
\footnotetext{``\textit{locally}'' refers to a measurement collected onboard, and thus can be used to estimate other agents' pose without needing to be transmitted.}

\addtocounter{footnote}{1}
\footnotetext{``\textit{instantaneous}'' implies the target's position is fully determined (i.e., observable) using only current measurements. As such, this approach does not require a dynamics model or measurements over a period of time. Furthermore, since estimation does not rely on integrating velocity or acceleration, the estimation should not drift/diverge over time.}

This paper's contributions are:
(1)~An in-depth analysis and modeling of the observed noise characteristics of 3D UWB ranging measurements (Section \ref{sec:extensions-3d}).
(2)~A 3D \textit{instantaneous} solution for UWB-based relative localization with minimal communication (i.e., only requires \textit{event-based} communication when a \textit{locally} monitored assumption is violated, which should occur infrequently or never). Our solution is formulated as a robust nonlinear least squares optimization with a learned measurement bias correction as a function of relative elevation,
improving our mean absolute position error by 19\% (Section \ref{sec:technical-approach}).
(3)~Hardware experimental results that demonstrate the merits of our proposed solution, \textbf{a mean APE and AHE of 0.24m and 9.5}\bm{$^\circ$} \textbf{respectively without \textit{continuously} transmitting measurements} (Section \ref{sec:experimental-results}).
(4)~Our datasets, over 200 hours of pairwise UWB range measurements with ground truth.\footnote{~\url{https://github.com/mit-acl/murp-datasets}}

\begin{figure*}[tp]
\centering
\hfill
\begin{subfigure}{.25\linewidth}
\includegraphics[height=2.5cm]{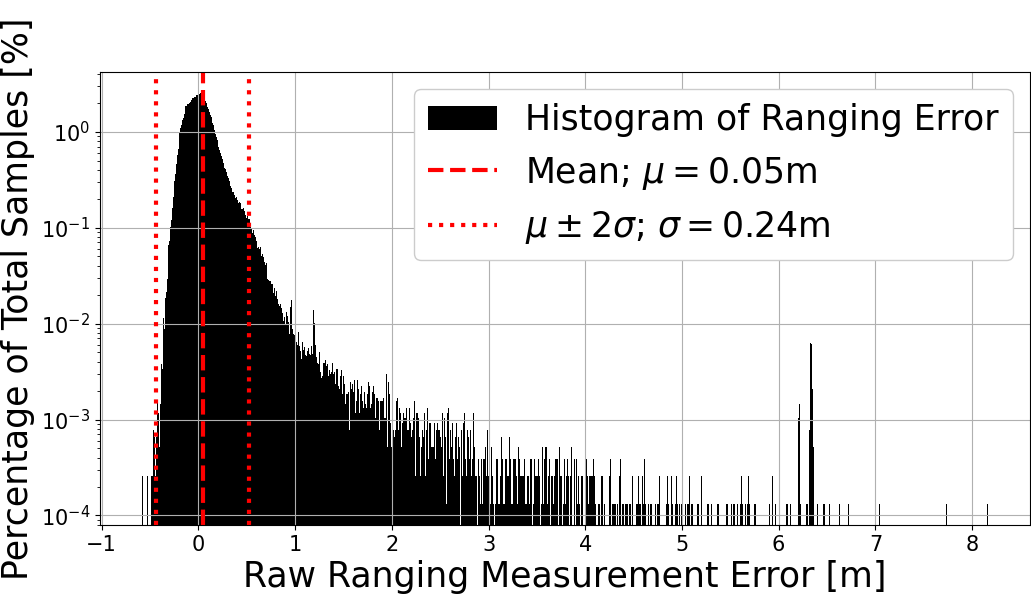}
\caption{}
\label{fig:long-tail}
\end{subfigure}
\hfill
\begin{subfigure}{.25\linewidth}
\includegraphics[height=2.5cm]{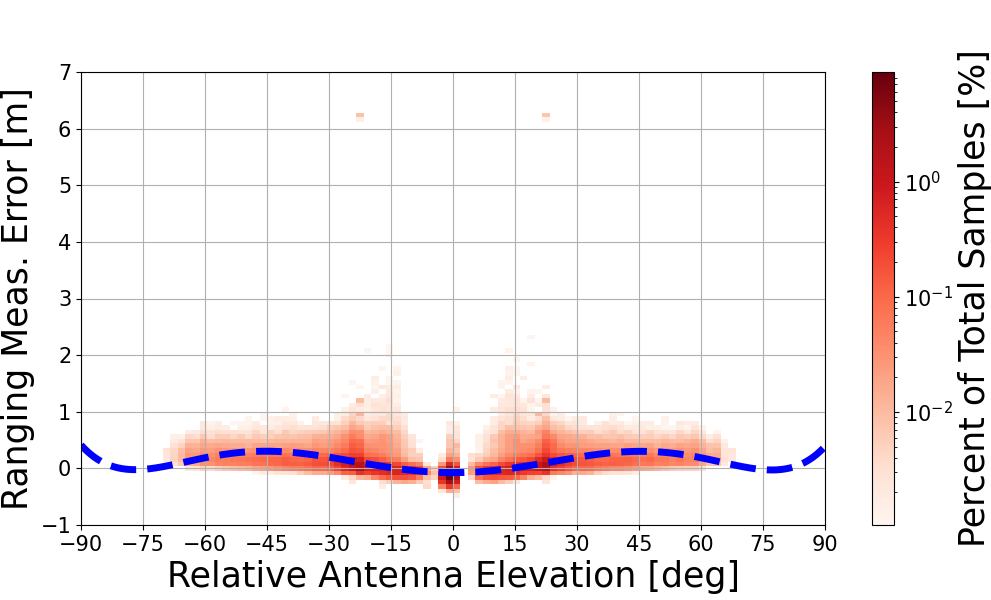}
\caption{}
\label{fig:noise-el}
\end{subfigure}
\hfill
\hspace{-.02\linewidth}
\begin{subfigure}{.25\linewidth}
\includegraphics[height=2.5cm]{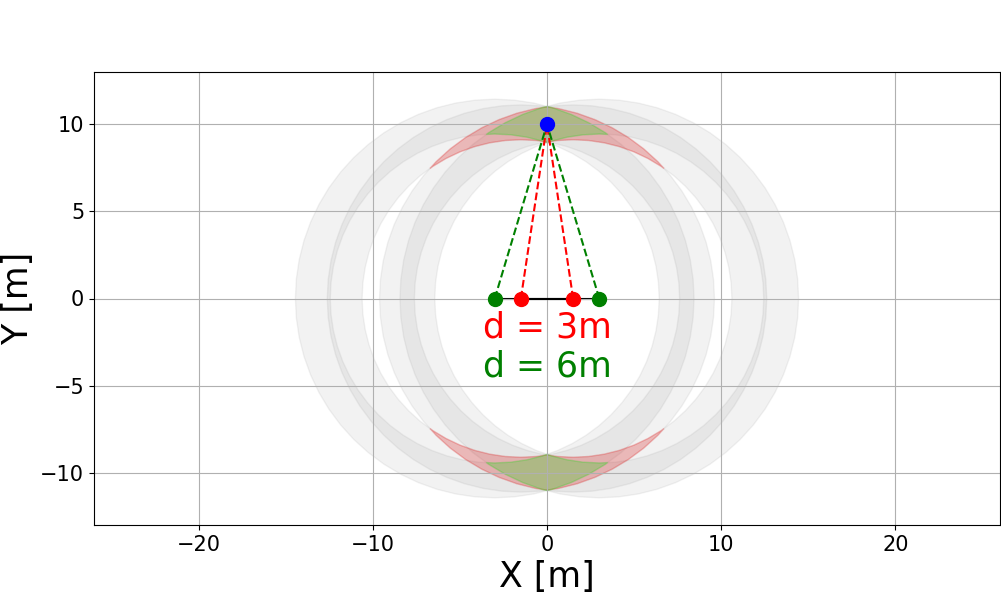}
\caption{}
\label{fig:dop}
\end{subfigure}
\hfill
\hspace{-.03\linewidth}
\begin{subfigure}{.25\linewidth}
\includegraphics[width=\linewidth]{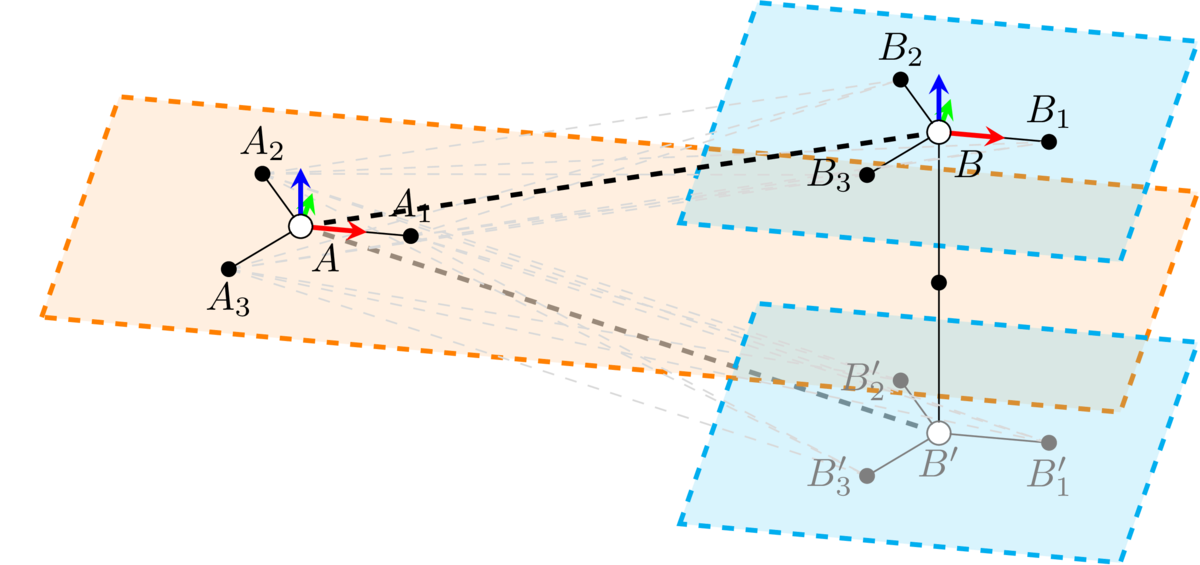}
\caption{}
\label{fig:3-ant-reflection}
\end{subfigure}
\hfill
\vspace*{-0.15cm}
\caption{Plots demonstrating the UWB noise and geometry characteristics outlined in Section \ref{sec:extensions-3d}.
\textbf{(a)} Histogram of our entire set of UWB measurements binned by range error. Demonstrates a non-zero mean and long tail (i.e., violates the zero mean Gaussian assumption that is typically used).
\textbf{(b)} Same data plotted as error with respect to relative elevation showing that the measurement error's mean and variance change significantly with relative elevation. The dotted blue line represents a learned 6-degree polynomial fit of measurement bias.
\textbf{(c)} Simple example demonstrating dilution of precision (DOP) in a 2D ranging scenario.
\textbf{(d)} Demonstrates how three (or more) ranging antennas within a single plane produce a pair of ambiguous solutions (i.e., if all antennas are in the base's plane $z=0$, while the target's $x$ and $y$ coordinates are fully observable, the target's altitude has an ambiguity between $\pm z$).}
\label{fig:extensions-to-3d}
\vspace{-4mm}
\end{figure*}

\section{Related Works}
\label{sec:related-works}

\newcommand{\YesGreen}{\cellcolor{green!25}Yes} %
\newcommand{\NoRed}{\cellcolor{red!25}No} %

\newcommand{\GreenBox}[1]{\cellcolor{green!25}#1}
\newcommand{\YellowBox}[1]{\cellcolor{yellow!25}#1}
\newcommand{\RedBox}[1]{\cellcolor{red!25}#1}

\newcommand{\CellHeight}{\\}

\begin{table}[!t]
\renewcommand{\arraystretch}{1.4}
\scriptsize
\caption{Comparison of relevant related works to highlight the capability gap this letter fills. Color emphasizes how a work fits in with our goals. Specifically, \textit{green} indicates our needs are met, \textit{yellow} indicates our needs are partially met, and \textit{red} indicates our needs are not met.
}
\label{tab:related-works}
\begin{centering}
\resizebox{\columnwidth}{!}{
\begin{tabular}{
| c |
>{\centering\arraybackslash}m{0.25\columnwidth}  ||
>{\centering\arraybackslash}m{0.18\columnwidth} 
>{\centering\arraybackslash}m{0.28\columnwidth}
>{\centering\arraybackslash}m{0.30\columnwidth} ||
}
\toprule 
\textbf{2D / 3D} &
\textbf{Related Work} & 
\textbf{\# UWB ~Per Agent} &
\textbf{UWB Noise Model} & 
\textbf{Data Exchanged} %
\\
\midrule
\multirow{8}{*}{\rotatebox[origin=c]{90}{\Large 2D}}
& Guo et al. \qquad\qquad (2020) \cite{guo_ultra-wideband_2020}
& \RedBox{Single}
& \YellowBox{Bounded Error Assumption}
& \RedBox{UWB + Velocity}
\CellHeight
& Cao et al. \qquad\qquad (2021) \cite{DBLP:conf/iros/Cao0YANMT21}
& \GreenBox{Many (4)}
& \YellowBox{NLLS}
& \RedBox{UWB + Odometry}
\CellHeight
& Zheng et al. \qquad\qquad (2022) \cite{zheng_uwb-vio_2022}
& \GreenBox{Many (2 or 4)}
& \YellowBox{ZM Gaussian + NLOS Rejection}
& \RedBox{UWB + VIO}
\CellHeight
& Zhang et al. \qquad\qquad (2023) \cite{zhang_range-aided_2023}
& \GreenBox{Many (2 or 4)}
& \YellowBox{ZM Gaussian + Outlier Rejection}
& \RedBox{UWB + Point Clouds + Odom + Keyframes}
\CellHeight
& \textbf{[Our Prev. Work]} \qquad\qquad (2022) \cite{DBLP:conf/iros/FishbergH22}
& \GreenBox{Many (4)}
& \GreenBox{NLLS + Data Informed Sensor Model}
& \GreenBox{UWB}
\CellHeight
\midrule
\multirow{10}{*}{\rotatebox[origin=c]{90}{\Large 3D}}
& Xu et al. \qquad\qquad (2020) \cite{DBLP:conf/icra/XuWZQS20}
& \RedBox{Single}
& \RedBox{ZM Gaussian}
& \RedBox{UWB + VIO + Visual Tracks}
\CellHeight
& Qi et al. \qquad\qquad\qquad (2020) \cite{qi_cooperative_2020}
& \RedBox{Single}
& \RedBox{ZM Gaussian}
& \RedBox{UWB + IMU + GPS Heading}
\CellHeight
& Xianjia et al. \qquad\qquad (2021) \cite{xianjia_cooperative_2021}
& \GreenBox{Many (2 or 4)}
& \YellowBox{NLLS}
& \GreenBox{UWB}
\CellHeight
& Xu et al. \qquad\qquad (2022) \cite{DBLP:journals/corr/abs-2103-04131}
& \RedBox{Single}
& \YellowBox{ZM Gaussian + Occlusion Rejection}
& \RedBox{UWB + VIO + Visual Tracks}
\CellHeight
& Xun et al. \qquad\qquad (2023) \cite{xun_crepes_2023}
& \RedBox{Single}
& \RedBox{ZM Gaussian}
& \RedBox{UWB + IMU + IR Visual Tracks}
\CellHeight
& \textbf{[This Work]} \qquad\qquad (2024)
& \GreenBox{Many (4 or 6)}
& \GreenBox{NLLS + Data Informed Sensor Model}
& \GreenBox{UWB + \textit{Event-Based} Assumption Violations}
\CellHeight
\bottomrule
\end{tabular}
}
\end{centering}
\vspace{-6mm}
\end{table}

UWB technology can be leveraged by mobile robotics in several orthogonal ways. This work uses UWB for infrastructure-free inter-agent relative measurements -- thus, works involving GPS position, static UWB anchors, %
UWB radar, or measurements done by direct waveform analysis are out of scope of this letter \cite{goudar_continuous-time_2023,mohanty2022precise,wen2020multi} (see survey paper \cite{DBLP:conf/meco/YuLQHW21} or our previous work \cite{DBLP:conf/iros/FishbergH22} for a broader review of UWB in robotics). Table \ref{tab:related-works} highlights the unique capability gap this work fills with respect to the recent related literature.
While all works in Table \ref{tab:related-works} use UWB for inter-agent ranging, it is challenging to compare numerical results given the vastly different operating assumptions and priorities -- specifically access to additional sensors (e.g., cameras, LiDAR, etc.), the size of agents with respect to their environment (see Section \ref{sec:trilateration-geometry}), and overall communication model (e.g., unlimited vs constrained communication) all significantly change performance and scalability. To contextualize this work, we discuss \cite{zhang_range-aided_2023} and \cite{DBLP:journals/corr/abs-2103-04131} since they are the most complete and comprehensive 2D and 3D UWB systems respectively, \cite{xun_crepes_2023} since it has a unique compact hardware solution, and \cite{xianjia_cooperative_2021} since it is the most directly comparable to this work.

Zhang et al.~\cite{zhang_range-aided_2023} presents a centralized 2D range-aided cooperative localization and consistent reconstruction system, merging a tightly coupled visual odometry and multi-tag ranging front-end with a pose graph optimization (PGO) back-end. This work differs from ours by being 2D and relying on \textit{continuous} transmission of point cloud, odometry, and keyframe information.
Omni-swarm \cite{DBLP:journals/corr/abs-2103-04131} is a decentralized 3D swarm estimation system. This work differs from ours by relying primarily on visual tracks of neighboring agents from each agent's omnidirectional camera -- omitting UWB measurements only degrades their estimation by approximately \meas{0.01m}. Additionally, each agent has only a single UWB tag and requires \textit{continuously} sharing measurements between all pairs of agents, scaling poorly with larger swarms.
CREPES \cite{xun_crepes_2023} presents a custom compact hardware module, which tightly couples an IMU, UWB, IR LEDs, and IR fish-eye camera. The measurements are fused into 6-DoF relative pose via a centralized error-state Kalman filter (ESKF) and PGO, while also having the ability to provide an \textit{instantaneous} estimate from a single frame of sensor measurements. This work differs from ours by relying heavily on visual tracks over UWB (similar to \cite{DBLP:journals/corr/abs-2103-04131}). Additionally, the system relies on \textit{continuously} transmitted measurements. Thus the state of the art approaches in \cite{zhang_range-aided_2023,DBLP:journals/corr/abs-2103-04131,xun_crepes_2023} achieve APE and AHE on the order of 0.10m and $1^\circ$ respectively, but require additional sensors that must \textit{continuously} transmit measurements, making them scale poorly to scenarios with larger swarm sizes or reduced communication throughput. Furthermore, these required additional sensors (e.g., LiDAR, specialized cameras, etc.) can cost thousands of dollars per agent, whereas UWB sensors (even multiple per robot) can cost $<$\$100 per agent.

Xianjia et al.~\cite{xianjia_cooperative_2021} is the most similar to our work (i.e., 3D environment, multiple UWB tags per agent, and using only UWB measurements). When their agents have similar antenna baselines to our work, they achieve a mean $xy$-positional error of approximately \meas{0.40m}, \meas{0.65m}, and \meas{0.85m} for simulated \meas{8m}$\times$\meas{8m} flights at various fixed altitudes. Furthermore, for real flights the mean $xy$-positional error becomes approximately \meas{1m} with a variable altitude and a (beneficial) larger baseline.
By comparison, although our experimental environment differs, we achieve a mean APE and AHE of \meas{0.24m} and $9.5^\circ$ respectively with real experiments operating in a \meas{10m}$\times$\meas{10m} mocap space. 
Ref.~\cite{xianjia_cooperative_2021} differs from our approach in two important ways: (1) the authors model UWB ranging error as a zero mean Gaussian with a \meas{0.10m} standard deviation, which does not reflect real data (see Section \ref{sec:characterizing-noise}); and (2) we define an explicit communication protocol that allows us to address several capability gaps without \textit{continuously} transmitting measurements.
Specifically, by constraining altitude/roll/pitch and improving our UWB noise model, we can better address UWB noise (see Section \ref{sec:characterizing-noise}) and an observability ambiguity 
(see Section \ref{sec:trilateration-geometry} and Figure \ref{fig:3-ant-reflection}). See Section \ref{sec:system-design} for more details.

\begin{table}[t]
\caption{Theoretical standard deviation of absolute position error between two of our UAVs (illustrated in Figure \ref{fig:system-diagram} and Section \ref{sec:uav-experiments}). Computation uses a UWB ranging uncertainty of $\sigma = 0.24$m (see Figure \ref{fig:long-tail}).
Altitude assumptions are the standard deviation of the target's assumed altitude when \textit{local only}, and $\pm0.04$m (the advertised precision of the downward LiDAR) when \textit{continuously sharing} measurements.
Since the system is nonlinear, table columns show evaluation at several relative altitudes. Table rows show how different relative altitude assumptions affect uncertainty.
Results demonstrate the advantage of relative altitude assumptions and shows how said assumptions enable comparable uncertainty with only \textit{locally} collected measurements vs systems that \textit{continuously} share altitude measurements.
See Section \ref{sec:system-design} for additional discussion.
}
\resizebox{\linewidth}{!}{%
\begin{tabular}{|c|c|c||c|c|c|c|c|c|}
\hline
\multicolumn{3}{|l||}{} & \multicolumn{6}{c|}{Evaluated at rel. position $[5\meas{m}, 0\meas{m}, {\Delta z}]$}                                                                                                                                                                                                                                                    \\ %
\multicolumn{3}{|l||}{\multirow{-2}{*}{\textbf{StdDev of Abs Pos Error [m]}}} %
& \multicolumn{6}{c|}{where $\Delta z$ equals:} \\
\hline
\textbf{name} & \textbf{local only} & \textbf{alt. assumpt.} & \textbf{0.0m} & \textbf{1.0m} & \textbf{2.5m} & \textbf{5.0m} & \textbf{10.0m} & \textbf{25.0m} \\
\hline \hline
\texttt{no\_assumpt} & \checkmark & n/a & $\infty$ & 9.98 & 2.05 & 1.35 & 2.81 & 14.38 \\
\texttt{extr\_conserv} & \checkmark & $\pm 2.00$m & 4.02 & 2.89 & 1.43 & 1.21 & 2.66 & 13.73 \\
\texttt{very\_conserv} & \checkmark & $\pm 1.00$m & 1.08 & 1.00 & 0.84 & 1.00 & 2.38 & 12.48 \\
\texttt{conserv} & \checkmark & $\pm 0.20$m & 0.39 & 0.41 & 0.49 & 0.79 & 1.96 & 10.21 \\
\rowcolor[HTML]{C0C0C0}
\texttt{proposed} & \checkmark & $\pm 0.10$m & 0.39 & 0.41 & 0.49 & 0.78 & 1.96 & 10.17 \\
\texttt{cont\_shared} &  & $\pm 0.04$m & 0.39 & 0.41 & 0.49 & 0.78 & 1.96 & 10.16 \\ \hline
\end{tabular}
}
\label{tab:theoretical-stdev}
\vspace{-6mm}
\end{table}

\section{Extensions to 3D Environments}
\label{sec:extensions-3d}

The following subsections motivate our hardware and algorithmic choices for operating in 3D (Section \ref{sec:system-design}). We first characterize the UWB ranging error of the Nooploop LinkTrack P-B \cite{hongli_linktrack_nodate} from a set of initial experiments\footnotemark\ (Section \ref{sec:characterizing-noise}), and then provide the observability and error properties of trilateration-based localization (Section \ref{sec:trilateration-geometry}).

\subsection{Characterizing UWB Noise}
\label{sec:characterizing-noise}

\bns
\textbf{Long Tail Distributions:}
Contrary to common noise assumptions in Table \ref{tab:related-works}, individual UWB ranging errors appear neither zero mean nor Gaussian. Instead, error distributions have long positive tails (see Figure \ref{fig:long-tail}). A trend towards positive bias can be attributed to the many ways positive ranging error can be introduced to a UWB measurement (e.g., multipath or a change of propagation medium).
Furthermore, installing SMA cables between RF devices and antennas adds consistent positive bias to all measurements.

\footnotetext{While these effects should generalize to other UWB ranging devices, the ``black box'' nature of commercial sensors means the exact manifestations could differ between products. Most works in Table \ref{tab:related-works} use a Nooploop LinkTrack P model \cite{hongli_linktrack_nodate}, except for \cite{guo_ultra-wideband_2020} (PulsON 440) and \cite{xianjia_cooperative_2021} (unspecified). Thus, we expect similar noise characteristics among those related works.}

\bs
\textbf{Sensitivity to Obstruction:}
Antenna obstruction (i.e., NLOS conditions) introduces positive bias and increased variance to collected ranging measurements (see Figure \ref{fig:obstruction}). While the UWB protocol is resilient to multipath and NLOS, UWB ranging is not completely absolved of these concerns. Furthermore, auxiliary metrics, like RSSI, do not appear to meaningfully indicate an obstructed measurement. Ref.~\cite{zheng_uwb-vio_2022,zhang_range-aided_2023,DBLP:conf/iros/FishbergH22,DBLP:journals/corr/abs-2103-04131} address this by rejecting suspect measurements, whether detected via statistical tests, robust loss functions, or hardcoded rejection criteria.

\bs
\textbf{Dependence on Relative Pose:}
Measurement noise is dependent on relative pose between antennas. The interplay of antenna attenuation patterns is a core concern of RF designers, but gets understandably overlooked by many end users. The LinkTrack P series comes equipped with a standard dipole antenna which we would expect to produce the behavior observed in Figure \ref{fig:noise-el} when aligned upright \cite{hongli_linktrack_nodate} -- specifically, approximately uniform performance within an $xy$-plane (i.e., varying azimuth) and degraded performance outside the ground plane $z=0$ (i.e., non-zero elevation). We note that both the mean bias and variance change with elevation. A relationship to elevation is noted in \cite{DBLP:journals/corr/abs-2103-04131}, although it is attributed to NLOS conditions similar to Figure \ref{fig:obstruction}, but \cite{DBLP:journals/corr/abs-2103-04131} chooses to address this by simply omitting range measurements with more than $37^\circ$ relative elevation (feasible only because of their reliance on other measurements).

\subsection{Geometry of Ranging-based Pose Estimation}
\label{sec:trilateration-geometry}

\bns
\textbf{Observability of Position:}
In 3D trilateration, a minimum of 4 non-coplanar antennas are required to uniquely determine a target's position \cite{lee1973accuracy}. That being said, 3 or more planar antennas produce only a pair of solutions that are a reflection across the antenna plane (see Figure \ref{fig:3-ant-reflection}). %
This ambiguity can be resolved with a simple $z$ measurement (i.e., altitude).

\bs
\textbf{Observability of Pose:}
Common in camera-robot registration,
Horn's Method is a closed-form solution for finding the pose between two Cartesian coordinate systems from a set of corresponding point pairs \cite{horn_closed-form_1987}.
Specifically, given exactly 3 planar points in two frames with known correspondences, Horn's Method fully specifies the 3D translation and orientation. Thus, by using trilateration to measure 3 points in a known configuration on our target, we can recover full pose.

\bs
\textbf{Dilution of Precision:}
Most commonly associated with GPS, dilution of precision (DOP) is a trilateration sensitivity analysis for quantifying how 1D ranging errors propagate to 2D/3D point estimation error \cite{richard_b_langley_dilution_1999}. Specifically, given trilateration's inherent nonlinearity, even identically noisy 1D range measurements can produce significantly different 2D/3D positional uncertainties -- this is entirely determined by the given base station and target geometry (Figure \ref{fig:dop}). In general we note that increasing the ``baseline" distance between base stations (i.e., the antenna separation on a single vehicle), or having a target closer to the base stations (i.e., decreasing distance between two vehicles), reduces trilateration's 2D/3D uncertainty. DOP also accounts for the improved accuracy seen by \cite{xianjia_cooperative_2021} for configurations with larger baselines. Thus, the accuracy of trilateration systems with vastly different baselines or operating ranges can be misleading. See \cite{zheng_uwb-vio_2022,zhang_range-aided_2023} for a detailed discussion of DOP as it pertains to 2D UWB localization.

\subsection{System Design}
\label{sec:system-design}

\begin{figure}
\centering
\hfill
\begin{subfigure}{.49\linewidth}
\includegraphics[width=\linewidth]{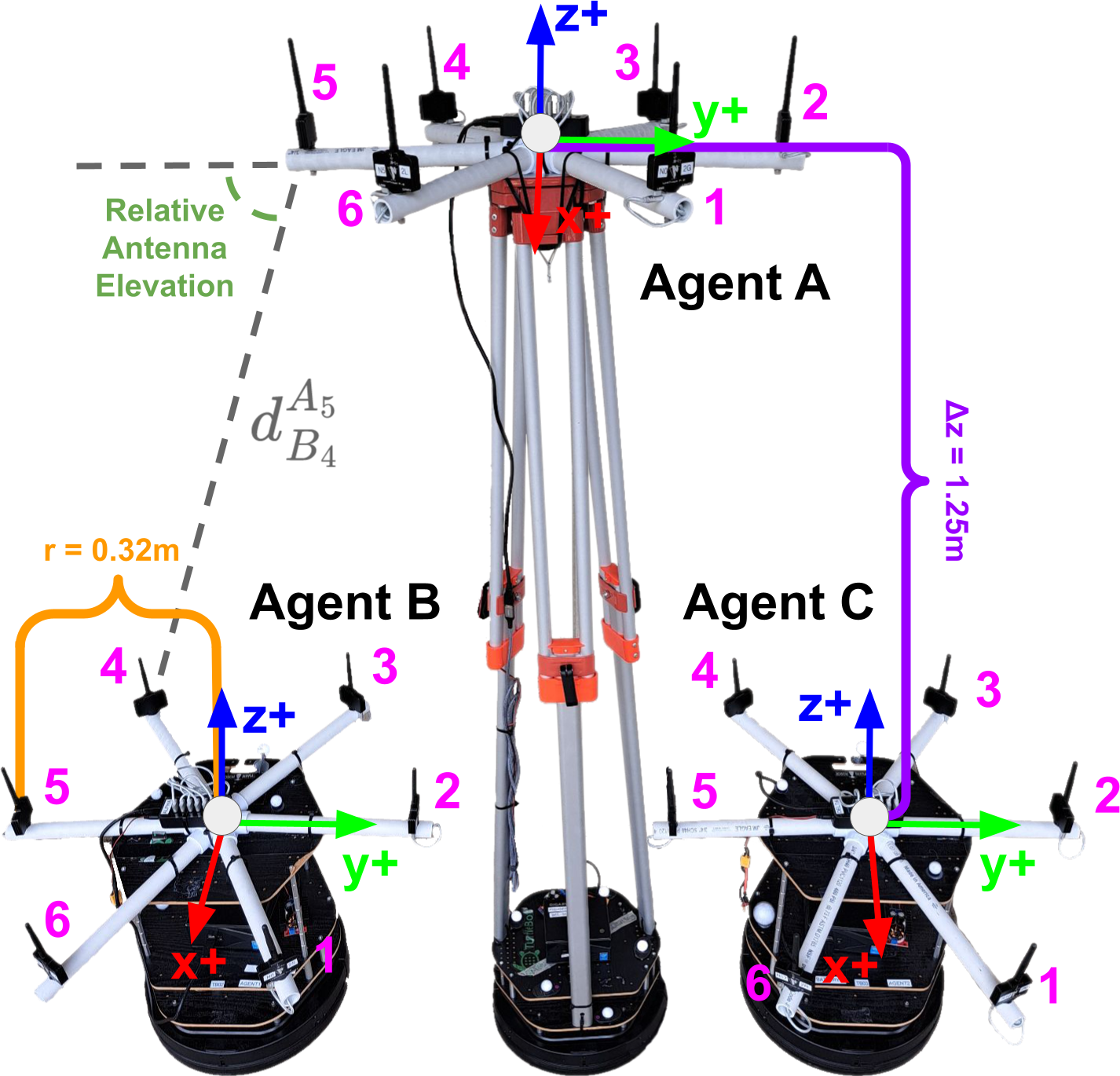}
\caption{}
\label{fig:ugv-system-diagram}
\end{subfigure}
\hfill
\begin{subfigure}{.49\linewidth}
\includegraphics[width=\linewidth]{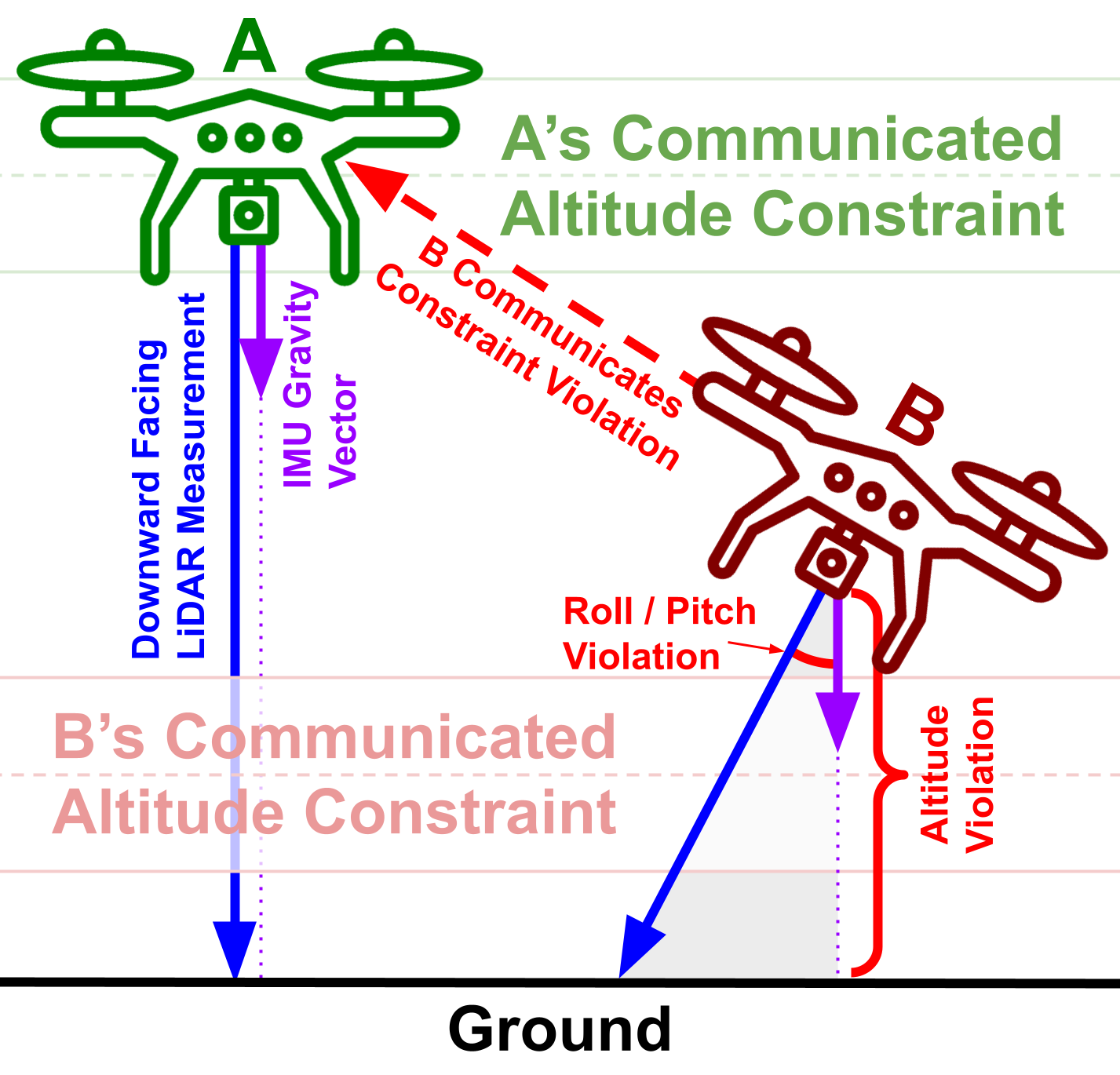}
\caption{}
\label{fig:altitude-violation}
\end{subfigure}
\hfill
\label{fig:experimental-diagrams}
\vspace*{-.10cm}
\caption{\textbf{(a)} Annotated diagram of three UGV agents used in Section \ref{sec:ugv-experiments}. UGV agents are designed with the same baseline as the UAV in Figure \ref{fig:system-diagram}, making them comparable surrogates to the UAV and its experiments in Section \ref{sec:uav-experiments}. \textbf{(b)} Diagram demonstrating agent $\B$ \textit{locally} detecting violations in its \textit{a priori} constraints, triggering an \textit{event-based} communication. Other agents will exclude $B$ from their relative pose estimation until $B$ notifies the swarm of restored constraint status or provides a new constraint envelop.}
\vspace{-6mm}
\end{figure}

Based on the noise characteristics (Section \ref{sec:characterizing-noise}) and underlying geometry (Section \ref{sec:trilateration-geometry}), we propose the system design shown in Figure \ref{fig:system-diagram} for our multirotor drones. Specifically, each agent is equipped with 6 ranging antennas (attached \meas{0.31m} from the center at the end of each propeller arm) to maximize the baselines and improve DOP geometry. Since there are 6 coplanar antennas, this is an over-constrained problem that still has a $\pm z$ ambiguity across the $z=0$ plane. Although an additional nonplanar antenna could be added to each agent, it is challenging to get a large $z$-baseline without impacting the flight characteristics (i.e., heavy/awkward configuration). Instead, we note that altitude/roll/pitch can be measured within the flat-ground world frame \textit{locally} via altimeter or downward facing LiDAR and onboard IMU.\footnote{Altitude measurements are direct and should not drift. Substantial drift in roll/pitch w.r.t. gravity would impede basic flight operations before meaningfully impacting our estimation, making it negligible for our system.} In many drone applications, it is common for multirotors to maintain roll/pitch near $0^\circ$ and a constant altitude. Thus, we can specify a minimalist communication protocol that shares these intended constraints \textit{a priori} and then monitors them \textit{locally} to ensure they are satisfied. \textit{Event-based} communication would then only occur if a constraint is \textit{locally} detected to have been violated/changed (Figure \ref{fig:altitude-violation}).

Table \ref{tab:theoretical-stdev} summarizes the theoretical uncertainty of the proposed system and demonstrates the value of adding relative altitude assumptions to a communication constrained system.\footnote{See explanation and code in the GitHub repository for more information on how the theoretical values in Table \ref{tab:theoretical-stdev} were computed.}
Although the assumed altitude bounds should vary with the platform, we see even \textit{overly} conservative assumptions (e.g., $\pm 2.0$m) still greatly improves performance over no assumption (\texttt{no\_assumpt} vs \texttt{extr\_conserv} or \texttt{very\_conserv}). This is especially true for level or near-level flight formations (i.e., unobservablility at $\Delta z = 0$ and high sensitivity nearby), a common configuration for many UAV systems.
By comparing the positional uncertainty of our proposed assumption (i.e., \texttt{proposed}, experimentally selected based on Figure \ref{fig:altitude}), a twice as lenient assumption (i.e., \texttt{conserv}), or \textit{continuously} communicated altitude measurements (i.e., \texttt{cont\_shared}, using the advertised downward LiDAR accuracy of $\pm 0.04$m) we see our minimal communication protocol does not substantially degrade uncertainty while using only \textit{locally} collected measurements.

\section{Technical Approach}
\label{sec:technical-approach}

\begin{algorithm}[tp]
\small
\caption{\footnotesize Procedure for robot $\A \in \R$, a member of robot swarm $\R$, to generate instantaneous relative poses w.r.t. all other agents in swarm $B \in (\R \setminus \{\A\})$. 
$A$ communicates violated assumptions with $\texttt{transmit}(A)$. $A$ \textit{locally} tracks which agents have communicated violated assumptions, and checks $B$'s last received status with $\texttt{invalid}(B)$.}
\label{alg:procedure}
\begin{algorithmic}[1]

\algrenewcommand\algorithmicrequire{\textbf{Input:}}
\algrenewcommand\algorithmicensure{\textbf{Output:}}

\Require $\R$, $\A \in \R$, $\{\I_R ~\forall~ R \in \R\}$, $\L$, $\dn\AiBj$
\Ensure $\{~\TAB ~\forall~ B \in (\R \setminus \{\A\}) \text{ s.t. } \neg \texttt{invalid}(B) ~\}$

\Function{LocalProcedure}{}

\State $\envelm_\A \gets $ \textit{local} measurements of $\A$ w.r.t. $\w$
\If{$\neg (|\envelc_\A - \envelm_\A| \leq \envelt_\A)$}
\State $\texttt{transmit}(\A)$
\EndIf

\State $\mathcal{T} \gets \{ \}$
\For{$\B \in (\R \setminus \{\A\}) \text{ s.t. } \neg\texttt{invalid}(B)$} %

\State $\dvm^\A_\B \gets $ \textit{local} measurements of $\B$ w.r.t. $\A$
\State $\tup{\zc\AB,\rollc\AB,\pitchc\AB} \gets \envelc\wB - \envelc\wA$
\State $\tup{\xAB,\yAB,\yawAB} \gets $ optimize Eq. \ref{eq:opt-f}
\State $\TAB \gets \T(\xAB,\yAB,\zc\AB,\rollc\AB,\pitchc\AB,\yawAB)$
\State $\mathcal{T} \gets \mathcal{T} \cup \{\TAB\}$

\EndFor

\State \Return $\mathcal{T}$

\EndFunction

\end{algorithmic}
\end{algorithm}

\subsection{Pose Parameterization}
Given a set of 3D reference frames $\F$, consider any pair of frames $\a \in \F$ and $\b \in \F$. Let $\Tab \in \SE{3}$ be the relative pose from frame $\a$ to frame $\b$. %
We note $\Tab$ is a 6-DoF value that can be equivalently parameterized as the 2-tuple $\langle\Rotab, \transab\rangle$ or 6-tuple $\langle\xab, \yab, \zab, \rollab, \pitchab, \yawab\rangle$, where $\Rotab \in \SO{3}$ is the relative rotation between $\a$ and $\b$ parameterized by relative roll $\rollab$, pitch $\pitchab$, and yaw $\yawab$ and $\transab \in \reals^3$ is a relative translation vector between $\a$ and $\b$ parameterized by relative $\xab$, $\yab$, and $\zab$.

\subsection{Local Robot Definitions}

Let there be some 3D world frame $\w$ defined with respect to the level ground and gravity vector (i.e., $z=0$ is the ``floor" and gravity faces down). Consider $\R$, a set of $N_\R$ robots operating in $\w$. Each robot $R \in \R$, has $N_R$ relative ranging antennas rigidly affixed to $R$'s body, where $R_k$ denotes $R$'s $k$th antenna. Additionally, for each $R$ let there be an \textit{a priori} and \textit{static} 3-tuple of known information $\I_R = \tup{\envelc_R, \envelt_R, \P_R}$, where:
\begin{itemize}
\item $\envelc_R = \begin{bmatrix} \zc\wR & \rollc\wR & \pitchc\wR \end{bmatrix}^\top$ is a vector of \textit{constant} global state constraints on $R$ with respect to world frame $\w$. In other words, $\zc\wR$ is the commanded altitude,
while $\rollc\wR$ and $\pitchc\wR$ are commanded relative roll and pitch.
\item $\envelt_R = \begin{bmatrix} \zt\wR & \rollt\wR & \pitcht\wR \end{bmatrix}^\top$ is a vector of \textit{constant} absolute (i.e., $\pm$) tolerances on %
global constraints $\envelc_R$.
\item $\mathcal{P}_R = \{\T^R_{R_1}, \T^R_{R_2}, \dots, \T^R_{R_{N_R}}\}$ is an ordered set of $N_R$ \textit{constant} relative poses from $R$'s body frame to $R$'s $k$th antenna's frame. From $\T^R_{R_k}$, we can succinctly denote the location of $R$'s $k$th antenna in $R$'s frame as point $\p^{R}_k$, where $\ph^{R}_k$ is the equivalent homogeneous point.
\end{itemize}
Since $\w$ is defined with respect to a level ground and gravity vector, the parameters $\tup{\z\wR, \roll\wR, \pitch\wR}$ -- unlike $\tup{\x\wR, \y\wR, \yaw\wR}$ -- are \textit{instantaneously} observable and \textit{directly} measurable from $R$ (e.g., via altimeter or downward facing LiDAR and onboard IMU). Thus, the directly measured values $\envelm_R = \begin{bmatrix} \zm\wR & \rollm\wR & \pitchm\wR \end{bmatrix}^\top$ can be continuously monitored by $R$ \textit{locally} (i.e., without relying on swarm communication). After an initial transmission of $\I_R$ to all agents $\R$, agent $R$ only needs to transmit again in the event of (1) wanting to change its commanded altitude/roll/pitch $\envelc_R$, or (2) it locally observes a violation of its previously communicated constraints, specifically:
\begin{equation}
\underbrace{|\envelc_R - \envelm_R|}_\text{element-wise abs deviation} \leq \underbrace{\envelt_R}_\text{constraint tolerance}
\end{equation}
where $|\cdot|$ and $\leq$ are performed element-wise. In other words, all agents $\R$ can treat $\envelc_R$ as \textit{known} states during estimation, unless $R$ communicates otherwise.

\subsection{Inter-Agent Robot Definitions}

Consider any arbitrary pair of robots $\A \in \mathcal{R}$ and $\B \in \mathcal{R}$. With a slight abuse of notation, let $A$ and $B$ also denote the robots respective reference frames, making $\TAB$ denote the relative pose between robots $A$ and $B$. From our known information tuples $\mathcal{I}_\A$ and $\mathcal{I}_\B$ we know each agent has $\NA$ and $\NB$ antennas at relative poses $\mathcal{P}_\A$ and $\mathcal{P}_\B$ respectively. This in turn allows us to define our observation model as just a function of $\TAB$. In other words, the distance between $\A$'s $i$th antenna and $\B$'s $j$th antenna is defined as:
\begin{equation}
d\AiBj(\TAB) \triangleq \lVert \TAB \ph^\B_j - \ph^\A_i \rVert_2
\end{equation}
where $d\AiBj(\TAB) \in \reals_{\geq0}$, as distances are non-negative, and $\lVert\cdot\rVert_2$ denotes the $L_2$ norm.
Similarly, let $\dm\AiBj$ be the current (i.e., most recent) noisy measurement between $\A$'s $i$th antenna and $\B$'s $j$th antenna as measured \textit{locally} by $\A$.\footnote{Note that while $\d\AiBj(\TAB) = \d^{\Bj}_{\Ai}(\T^{\B}_{\A})$ (i.e., the observation model is symmetric), since each agent \textit{locally} and \textit{separately} measures antenna distances, measurement noise makes it that generally $\dm\AiBj \neq \dm^\Bj_\Ai$.}
For convenience, $\dvm\AB \in \reals^{\NA\NB}$ denotes the stacked vector of all current pairwise $\dm\AiBj$ measurements.

\subsection{Noise Model Definition}
Let $\TAiBj$ denote the relative pose between $\A$'s $i$th antenna and $\B$'s $j$th antenna. Given a known $\P_\A$ and $\P_\B$, and an arbitrary $\TAB$, we have:
\begin{equation}
\TAiBj = (\T^\A_{\A_i})^{-1} \TAB \T^\B_{\B_j} = \T^{\A_i}_\A \TAB \T^\B_{\B_j}
\end{equation}
so that $\TAiBj$ is a function of $\TAB$ given known $\P_\A$ and $\P_\B$.
Let $\dn(\TAiBj) \in \reals$ be some provided (e.g., learned) sensor mean bias model as a function of $\TAiBj$ (i.e., antenna measurement bias is modeled as some function of the relative position and/or orientation between a pair of ranging antennas). We can then equivalently define:
\begin{equation}
\dn\AiBj(\TAB) \triangleq \dn(\T^{\A_i}_\A \TAB \T^\B_{\B_j})
\end{equation}
where $\dn\AiBj(\TAB) \in \reals$ is a state-dependent mean bias estimate for measurement $\dm\AiBj$.

\subsection{Optimization Definition}
\label{sec:opt}

Consider the following error residual function between the measured and expected distances for antenna pair $\Ai$ and $\Bj$ with respect to some $\TAB$, a relative pose between $\A$ and $\B$ (and our eventual decision variable):
\begin{equation}
\e\AiBj(\TAB) \triangleq \underbrace{\BigP{\dm\AiBj - \dn\AiBj(\TAB)}}_\text{bias adjusted measurement} - \underbrace{\d\AiBj(\TAB)}_\text{observation model}
\end{equation}
where $\dm\AiBj$ is the current measurements, $\dn\AiBj(\TAB)$ is a measurement bias correction function (current implementation in Section \ref{sec:ugv-experiments}), and $\d\AiBj(\TAB)$ the known observation model.

Using our error residual function $\e\AiBj(\TAB)$, and some loss function $\L$, we can calculate an instantaneous estimate of $\TAB$ by minimizing the sum of loss of all error residuals for all inter-agent antenna pairs with respect to $\TAB$. That is:
\begin{equation}
\min_{\TAB \in SE(3)} \sumi \sumj \L\BigP{\e\AiBj(\TAB)}
\end{equation}
Recall that $\TAB$ is a 6-DoF variable parameterized by $\tup{\xAB,\yAB,\zAB,\rollAB,\pitchAB,\yawAB}$. From known information tuples $\I_\A$ and $\I_\B$, we have $\envelc_\A$ and $\envelc_\B$, which provide us with constraints $\tup{\zc\wA, \rollc\wA, \pitchc\wA}$ and $\tup{\zc\wB, \rollc\wB, \pitchc\wB}$.
Furthermore, since these constraints are with respect to a common world frame $\w$, see that: $\zc\AB = \zc\wB - \zc\wA$, $\rollc\AB = \rollc\wB - \rollc\wA$, and $\pitchc\AB = \pitchc\wB - \pitchc\wA$. Thus, given $\envelc_A$ and $\envelc_B$, we have the following constrained optimization:
\begin{align}
\begin{split}
\min_{\TAB \in SE(3)} \quad& \sumi \sumj \L\BigP{\e\AiBj(\TAB)} \\
\text{s.t.} \quad & \zAB = \zc\AB = \zc\wB - \zc\wA \\
& \rollAB = \rollc\AB = \rollc\wB - \rollc\wA \\
& \pitchAB = \pitchc\AB = \pitchc\wB - \pitchc\wA
\end{split}
\label{eq:constraints}
\end{align}
which can then be simplified to an equivalent 3-DoF unconstrained optimization problem:
\begin{equation}
\scalemath{.85}{\min_{\substack{\xAB,\yAB \in \reals \\ \yawAB \in [-180^\circ,180^\circ]}} \sumi \sumj \L\BiggP{\e\AiBj\BigP{\T(\underbrace{\xAB,\yAB}_{\text{free}}, \underbrace{\zc\AB, \rollc\AB, \pitchc\AB}_{\text{constrained}}, \underbrace{\yawAB}_{\text{free}})}}}
\end{equation}
which can be rewritten as $f$ to clearly see the dependencies:
\begin{equation}
\label{eq:opt-f}
\min_{\substack{\xAB,\yAB \in \reals \\ \yawAB \in [-180^\circ,180^\circ]}} f(\xAB,\yAB,\yawAB ~|~ \dvm\AB, \I_\A, \I_\B, \L, \dn\AiBj)
\end{equation}
Using Equation \ref{eq:opt-f}, Algorithm \ref{alg:procedure} outlines a procedure independently followed by each agent $\A$ in a swarm $\R$ to individually estimate $\TAB$ between itself and all other agents $\B \in (\R \setminus \{A\})$. Observe that \textit{event-based} communication (i.e., \texttt{transmit}) only occurs on Line 4, when an agent \textit{locally} detects it has violated its \textit{a priori} constraints -- other agents use this information on Line 7 (i.e., \texttt{invalid}) to skip calculating $\TAB$ for agents with violated constraints.

\begin{table}[tp]
\caption{Evaluation of positional error in meters with different $z$-constraints. In practice, the second row (i.e., \textit{a priori} constraints) will be used, but the small difference between \texttt{z\_comm}, \texttt{z\_meas}, and \texttt{z\_true} shows these constraints are not a major source of error. See Section \ref{sec:uav-experiments}.}
\resizebox{\linewidth}{!}{%
\begin{tabular}{|>{\centering}m{.6cm}|c||c c c|c c c|c c c|}
\hline
\multicolumn{2}{|l||}{\multirow{2}{*}{\textbf{Abs Position Error [m]}}} & \multicolumn{9}{|c|}{Scenarios (Section \ref{sec:uav-experiments})} \\
\cline{3-11} \multicolumn{2}{|l||}{} & \multicolumn{3}{|c|}{Trial 1} & \multicolumn{3}{|c|}{Trial 2} & \multicolumn{3}{|c|}{Trial 3} \\
\cline{1-11} \textbf{color} & \textbf{alt constraint} & \textbf{Mean} & \textbf{Max} & \textbf{Std} & \textbf{Mean} & \textbf{Max} & \textbf{Std} & \textbf{Mean} & \textbf{Max} & \textbf{Std} \\
\hline \hline
 & \texttt{z\_free} & 1.17 & 2.83 & 1.06 & 1.04 & 1.53 & 0.24 & 0.45 & 0.94 & 0.19 \\
\rowcolor[gray]{.8} \textcolor{tab-blue}{$\blacksquare$} & \texttt{z\_comm} & 0.26 & 0.45 & 0.09 & 0.28 & 0.87 & 0.14 & 0.25 & 0.64 & 0.13 \\
\textcolor{tab-red}{$\blacksquare$} & \texttt{z\_meas} & 0.26 & 0.44 & 0.09 & 0.27 & 0.87 & 0.14 & 0.24 & 0.66 & 0.14 \\
\textcolor{tab-green}{$\blacksquare$} & \texttt{z\_true} & 0.25 & 0.44 & 0.10 & 0.26 & 0.87 & 0.15 & 0.22 & 0.64 & 0.14 \\
\hline
\end{tabular}%
}
\label{tab:flight}
\vspace{-4mm}
\end{table}

\begin{figure}[tp]
\centering
\hfill
\begin{subfigure}{\linewidth}
\includegraphics[width=\linewidth]{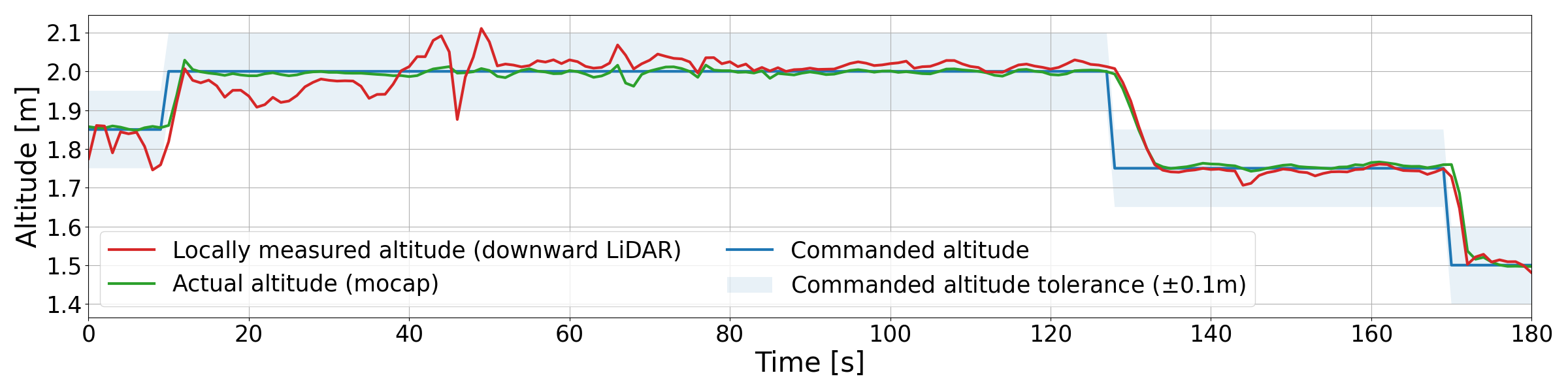}
\vspace{-.6cm}
\caption{Shows Table \ref{tab:flight}'s Trial 2 alongside the commanded, measured, and ground truth altitude, where the colors correspond with Table \ref{tab:flight}. While the TeraRanger Evo does not perfectly track ground truth, it stays within the expected commanded tolerance of $\pm 0.10$m, providing a reasonable \textit{local} measurement of altitude.}
\label{fig:altitude}
\end{subfigure}

\vspace{.2cm}
\begin{subfigure}{\linewidth}
\includegraphics[width=\linewidth]{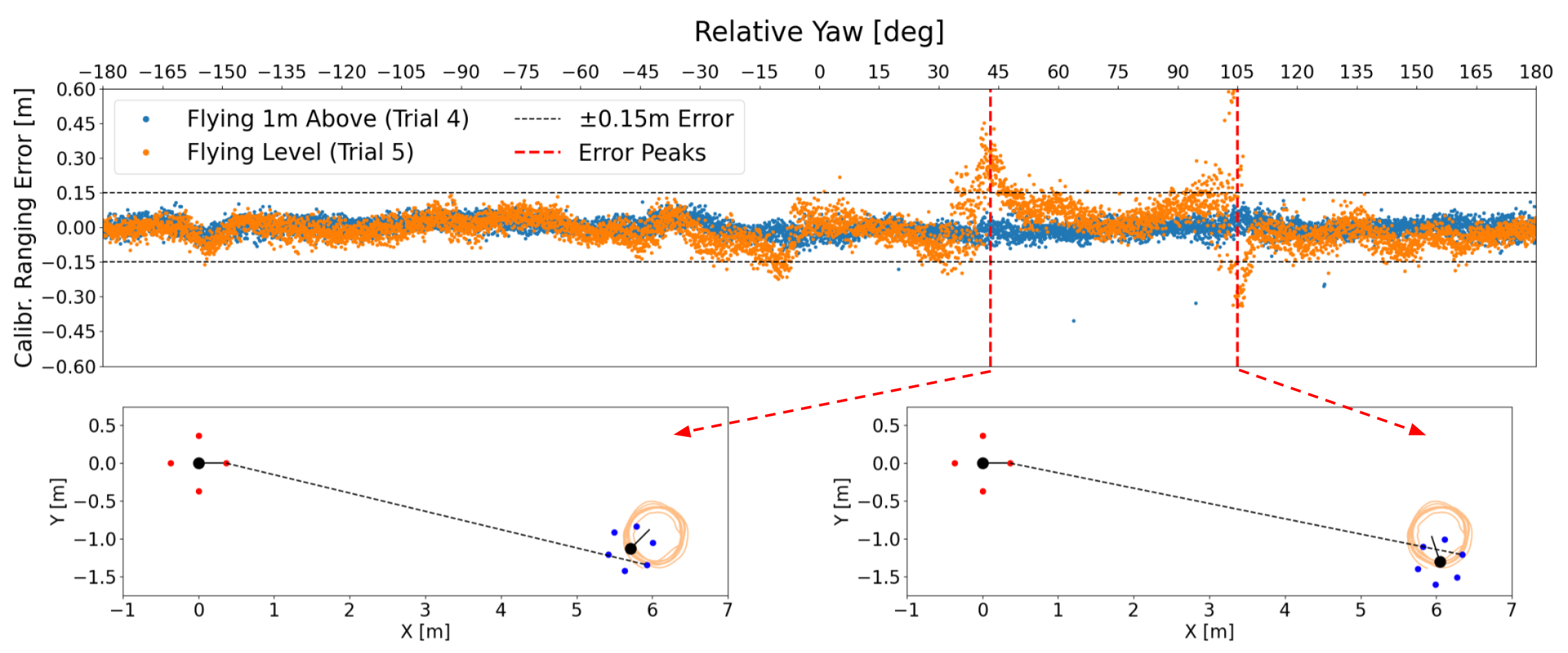}
\vspace{-.6cm}
\caption{Top plot shows calibrated ranging error (i.e., mean bias subtracted) with respect to relative yaw for the \textit{1m Above} and \textit{Level} flights (i.e., Trial 4 and 5 respectively). Observe the error of \textit{1m Above} flight remains consistent, while the \textit{Level} flight has anomalous error peaks, mostly within the $\pm 0.15$m range, corresponding to specific relative yaws. The lower left and right diagrams show the relative pose of the agents at the two most prominent peaks (i.e., with error greater than $0.45$m), where the path between antennas $\A_1$ and $\B_5$ is eclipsed by antennas $\B_3$ and $\B_1$ respectively. While this NLOS effect is consistent with our previous 2D work \cite{DBLP:conf/iros/FishbergH22}, we note the more complex 3D relationship resolves the occlusion with varying elevation.}
\label{fig:obstruction}
\end{subfigure}

\caption{Plots from the UAV flight experiments.}
\label{fig:flight-tests}
\vspace{-6mm}
\end{figure}

\begin{table*}[ht!]
\caption{Evaluation of positional error in meters between algorithms. The final row (red) represents the proposed approach. See Section \ref{sec:ugv-experiments}.}
\resizebox{\textwidth}{!}{%
\begin{tabular}{| >{\centering}m{.4cm} | >{\centering}m{1cm} >{\centering}m{1cm} >{\centering}m{1cm}||c c c|c c c|c c c|c c c|c c c|}
\hline
\multicolumn{4}{|l||}{\multirow{2}{*}{\textbf{Abs Position Error [m]}}} & \multicolumn{15}{|c|}{Experimental Scenarios (Section \ref{sec:ugv-experiments})} \\
\cline{5-19} \multicolumn{4}{|l||}{} & \multicolumn{3}{|c|}{Trial 1} & \multicolumn{3}{|c|}{Trial 2} & \multicolumn{3}{|c|}{Trial 3} & \multicolumn{3}{|c|}{Trial 4} & \multicolumn{3}{|c|}{Trial 5} \\
\hline {\tiny \textbf{color}} & {\tiny \textbf{el\_bias}} & {\tiny \textbf{z\_fixed}} & {\tiny \textbf{Huber}} & \textbf{Mean} & \textbf{Max} & \textbf{Std} & \textbf{Mean} & \textbf{Max} & \textbf{Std} & \textbf{Mean} & \textbf{Max} & \textbf{Std} & \textbf{Mean} & \textbf{Max} & \textbf{Std} & \textbf{Mean} & \textbf{Max} & \textbf{Std} \\
\hline \hline
\textcolor{tab-purple}{$\blacksquare$} &  &  &  & 2.21 & 4.40 & 1.17 & 0.89 & 4.50 & 0.88 & 2.37 & 4.82 & 1.13 & 2.26 & 4.58 & 1.12 & 2.77 & 4.32 & 0.59 \\
 & \checkmark &  &  & 2.55 & 4.95 & 0.64 & 0.89 & 4.46 & 0.97 & 1.32 & 3.94 & 0.95 & 1.33 & 3.97 & 0.97 & 1.55 & 3.71 & 0.90 \\
 & & \checkmark &  & 0.45 & 2.87 & 0.38 & 0.42 & 1.50 & 0.28 & 0.57 & 2.97 & 0.52 & 0.45 & 2.40 & 0.36 & 0.40 & 1.16 & 0.23 \\
 & \checkmark & \checkmark &  & 0.34 & 2.86 & 0.43 & 0.34 & 1.49 & 0.32 & 0.53 & 2.96 & 0.55 & 0.38 & 2.41 & 0.39 & 0.30 & 1.16 & 0.28 \\
\hline
\textcolor{tab-blue}{$\blacksquare$} & &  & \checkmark & 1.14 & 3.29 & 1.12 & 2.15 & 3.29 & 1.03 & 1.24 & 3.60 & 1.01 & 2.30 & 3.57 & 1.02 & 1.30 & 3.24 & 1.16 \\
\textcolor{tab-green}{$\blacksquare$} & \checkmark &  & \checkmark & 0.44 & 1.64 & 0.30 & 0.95 & 2.58 & 0.90 & 0.85 & 3.10 & 0.75 & 0.85 & 2.69 & 0.82 & 1.44 & 2.88 & 0.90 \\
\textcolor{tab-orange}{$\blacksquare$} & & \checkmark & \checkmark & 0.29 & 0.91 & \textbf{0.14} & 0.27 & 1.02 & \textbf{0.14} & 0.32 & 1.03 & \textbf{0.20} & 0.28 & 0.78 & \textbf{0.14} & 0.29 & 0.71 & \textbf{0.13} \\
\rowcolor[gray]{.8} \textcolor{tab-red}{$\blacksquare$} & \checkmark & \checkmark & \checkmark & \textbf{0.22} & \textbf{0.90} & 0.17 & \textbf{0.21} & \textbf{0.98} & 0.16 & \textbf{0.30} & \textbf{0.97} & 0.21 & \textbf{0.23} & \textbf{0.75} & 0.15 & \textbf{0.22} & \textbf{0.63} & 0.14 \\
\hline
\end{tabular}%
}
\label{tab:pos-err}

\caption{Evaluation of yaw error in degrees between algorithms. The final row (red) represents the proposed approach. See Section \ref{sec:ugv-experiments}.}
\resizebox{\textwidth}{!}{%
\begin{tabular}{| >{\centering}m{.4cm} | >{\centering}m{1cm} >{\centering}m{1cm} >{\centering}m{1cm}||c c c|c c c|c c c|c c c|c c c|c c c|c c c|c c c|c c c|c c c|}
\hline
\multicolumn{4}{|l||}{\multirow{2}{*}{\textbf{Abs Heading Error [deg]}}} & \multicolumn{15}{|c|}{Experimental Scenarios (Section \ref{sec:ugv-experiments})} \\
\cline{5-19} \multicolumn{4}{|l||}{} & \multicolumn{3}{|c|}{Trial 1} & \multicolumn{3}{|c|}{Trial 2} & \multicolumn{3}{|c|}{Trial 3} & \multicolumn{3}{|c|}{Trial 4} & \multicolumn{3}{|c|}{Trial 5} \\
\hline {\tiny \textbf{color}} & {\tiny \textbf{el\_bias}} & {\tiny \textbf{z\_fixed}} & {\tiny \textbf{Huber}} & \textbf{Mean} & \textbf{Max} & \textbf{Std} & \textbf{Mean} & \textbf{Max} & \textbf{Std} & \textbf{Mean} & \textbf{Max} & \textbf{Std} & \textbf{Mean} & \textbf{Max} & \textbf{Std} & \textbf{Mean} & \textbf{Max} & \textbf{Std} \\
\hline \hline
\textcolor{tab-purple}{$\blacksquare$} &  &  &  & 13.4 & 133.0 & 18.8 & 10.0 & 139.9 & 12.2 & 15.0 & 146.0 & 19.3 & 15.5 & 129.6 & 19.2 & 13.1 & 162.2 & 14.0 \\
 & \checkmark &  &  & 13.4 & 134.8 & 18.8 & 10.0 & 139.9 & \textbf{12.1} & 15.1 & 146.0 & 19.5 & 15.6 & 132.9 & 19.4 & 13.3 & 162.1 & 14.0 \\
 &  & \checkmark &  & 13.5 & 132.8 & 18.8 & 10.1 & 139.9 & \textbf{12.1} & 15.0 & 146.5 & 19.3 & 15.5 & 131.9 & 19.3 & 13.2 & 152.4 & 13.5 \\
 & \checkmark & \checkmark &  & 13.8 & 133.0 & 18.7 & 10.4 & \textbf{139.0} & 12.2 & 15.0 & 146.2 & 19.3 & 16.8 & 170.1 & 23.1 & 14.0 & 166.7 & 14.4 \\
\hline
\textcolor{tab-blue}{$\blacksquare$} &  &  & \checkmark & 8.0 & 47.5 & 5.5 & 7.7 & 177.7 & 12.8 & 8.5 & 57.5 & \textbf{6.1} & \textbf{10.1} & 111.4 & \textbf{10.7} & \textbf{10.4} & 47.5 & \textbf{7.6} \\
\textcolor{tab-green}{$\blacksquare$} & \checkmark &  & \checkmark & 7.8 & 47.7 & 5.7 & 7.8 & 177.3 & 12.8 & 8.5 & \textbf{57.3} & \textbf{6.1} & 10.3 & \textbf{111.2} & 11.0 & \textbf{10.4} & 46.5 & \textbf{7.6} \\
\textcolor{tab-orange}{$\blacksquare$} &  & \checkmark & \checkmark & \textbf{7.6} & \textbf{42.4} & \textbf{5.1} & \textbf{7.6} & 172.6 & 12.5 & \textbf{8.3} & 58.4 & \textbf{6.1} & 10.4 & 122.0 & 11.2 & 10.6 & 48.5 & \textbf{7.6} \\
\rowcolor[gray]{.8} \textcolor{tab-red}{$\blacksquare$} & \checkmark & \checkmark & \checkmark & 8.6 & 46.7 & 5.8 & 8.4 & 172.8 & 12.8 & 8.7 & 59.8 & 6.5 & 10.7 & 111.8 & 11.2 & 11.0 & \textbf{42.3} & 7.7 \\
\hline
\end{tabular}%
}
\label{tab:yaw-err}
\end{table*}

\begin{figure*}
\centering
\hfill
\begin{subfigure}{.49\textwidth}
\includegraphics[width=\linewidth]{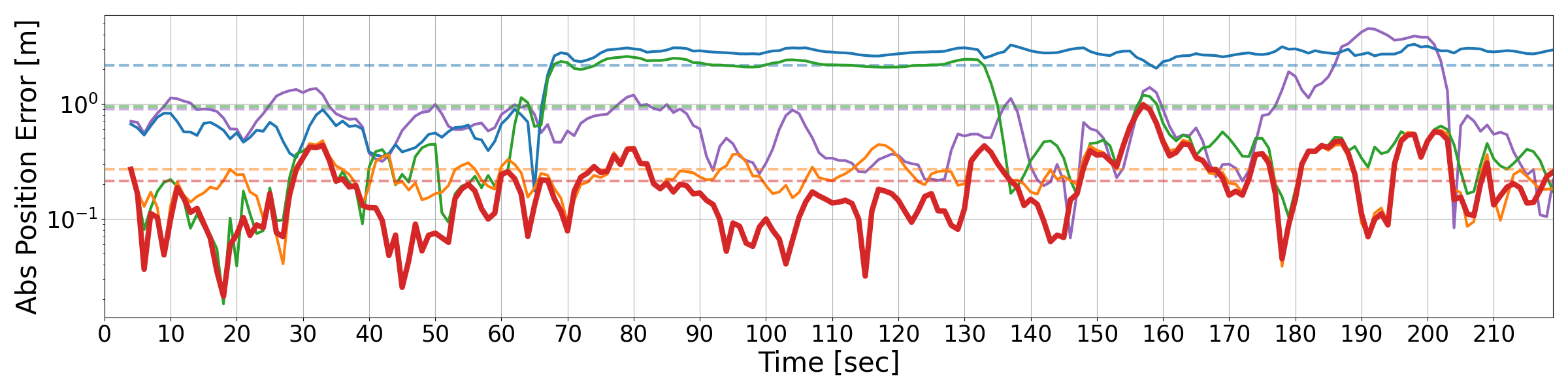}
\label{fig:ape-vs-time}
\end{subfigure}
\hfill
\begin{subfigure}{.49\textwidth}
\includegraphics[width=\linewidth]{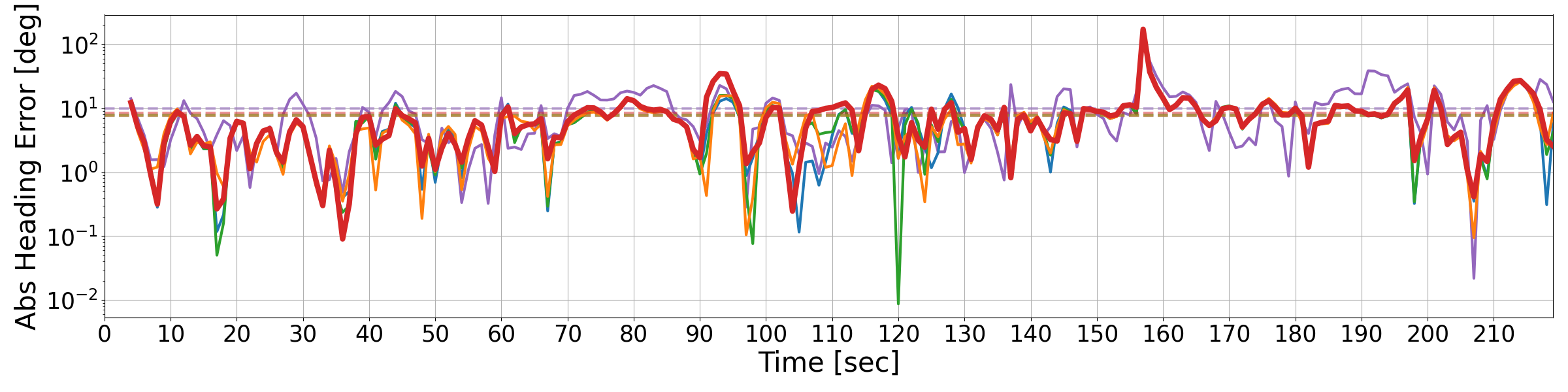}
\label{fig:ahe-vs-time}
\end{subfigure}
\hfill\null

\vspace{-5mm}
\caption{Error vs Time plots for Trial 2 in Table \ref{tab:pos-err} and \ref{tab:yaw-err} respectively. The line color corresponds with the algorithm's color column in the associated tables. Each colored dashed line represent that algorithm's overall mean in that trial. }
\label{fig:ugv-plots}
\vspace{-4mm}
\end{figure*}

\begin{figure}
\centering
\includegraphics[width=\linewidth]{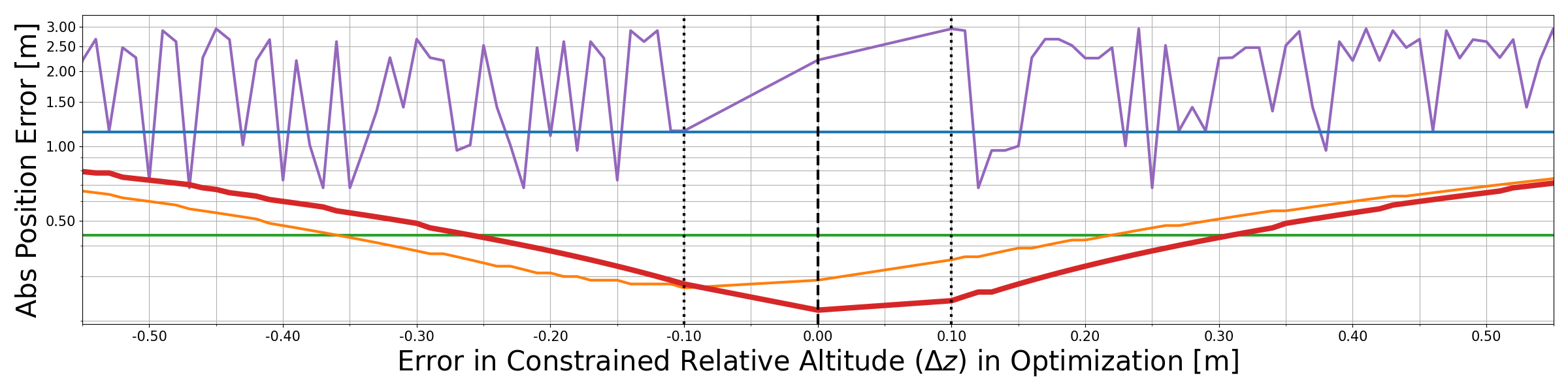}
\vspace*{-0.6cm}
\caption{Shows the corresponding absolute position error for Trial 1 when each approach (colors correspond with Table \ref{tab:pos-err}) is run with the incorrect altitude $z$-constraint $\zc\AB \triangleq -1.25\text{m} + \Delta z$, where $-1.25$m is the true relative altitude between $\A$ and $\B$ (Figure \ref{fig:ugv-system-diagram}) and $\Delta z$ is relative altitude error (i.e., the $x$-axis value).
The green/blue lines are level since $z$ is unconstrained in these methods, and thus not depended on the $x$-axis.
Although purple also has $z$ unconstrained, without Huber loss, it is hyper-sensitive to altitude initialization (i.e., $x$-axis value). Comparing our proposed (red) to the approach without \texttt{el\_bias} (orange), we see red outperforms orange even when operating with \textit{maximal} altitude error within the reasonable error tolerances (i.e., $\pm 0.10$m) demonstrated in Figure \ref{fig:altitude}.} %
\label{fig:pos-err-vs-alt-err}
\vspace{-6mm}
\end{figure}

\section{Experimental Results}
\label{sec:experimental-results}

We present two sets of experiments: UAV (Section \ref{sec:uav-experiments}) and UGV (Section \ref{sec:ugv-experiments}). 
We first verify the 3D noise properties, flight hardware, and show altitude/roll/pitch constraints can be sufficiently \textit{locally} monitored on a UAV.
Once this is verified, we attach surrogate UAVs at different altitudes to UGVs -- this allows us to efficiently collect large amounts of data to evaluate our approach.

\subsection{UAV Experiments}
\label{sec:uav-experiments}

\bns
\textbf{Hardware:}
Two agents, $\A$ and $\B$, were used. $\A$, a surrogate quadcopter, was a stationary Turtlebot2 UGV with 4 UWB sensors evenly distributed in the $xy$-plane \meas{0.37m} from the center (i.e., same robot in prior work \cite{DBLP:conf/iros/FishbergH22}). $\B$ is a custom hexrotor UAV equipped with 6 UWB sensors evenly distributed in the $xy$-plane \meas{0.31m} from the center (Figure \ref{fig:system-diagram}). Additionally, $\B$ is equipped with a downward facing TeraRanger Evo 15m LiDAR for altitude measurements \cite{fabre_teraranger_2021}.

\bs
\textbf{Trials:} Five experimental trials were conducted in a large \meas{10m}$\times$\meas{10m} mocap space. For all trials, $\A$ was placed on a table for additional height and kept stationary, while $\B$ flew around. In the first three trials, $\B$ took off and flew simple line patterns at different altitudes (see Table \ref{tab:flight} and Figure \ref{fig:altitude}). For the latter two trials, $\B$ flew tight circles in the center of the room -- first \textit{\meas{1m} Above} and then \textit{Level} with $\A$ (see Figure \ref{fig:obstruction}). Each trial had a max speed of \meas{1m/s} and lasted approximately \meas{3-5min} (i.e., the length of a battery charge).

\bs
\textbf{Discussion of Results:} Results are shown in Table \ref{tab:flight} and Figure \ref{fig:flight-tests}. The key takeaways are as follows:
(1)~Demonstrated UWB measurements were not corrupted by proximity to propeller motors (e.g., electromagnetic interference). This gives us confidence measurements in Section \ref{sec:ugv-experiments} are comparable.
(2)~Showed the 2D NLOS concerns of prior work \cite{DBLP:conf/iros/FishbergH22} are mitigated when level 3D agents are not in the same $z$-plane (Figure \ref{fig:obstruction}).
(3)~Confirmed the onboard downward facing LiDAR and IMU can \textit{locally} monitor altitude/roll/pitch accurately with only small variation from the mocap ground truth. Figure \ref{fig:altitude} shows data from Table \ref{tab:flight}'s Trial 2, demonstrating the TeraRanger Evo provides a reasonable \textit{local} measurement of altitude.
(4)~Confirmed varying within our \textit{a priori} tolerances are not a major source of APE error, supported by the small APE differences between the latter three rows of Table \ref{tab:flight}. Specifically, this table shows the results of the proposed algorithm
when $z$ is free (\texttt{z\_free}), or fixed to the commanded (\texttt{z\_comm}), LiDAR measured (\texttt{z\_meas}), or true (\texttt{z\_true}) value. 
Together, takeaways (3) and (4) show that hardware realistically allows for an altitude tolerance of $\pm 0.10$m, meaning under realistic flight conditions, our minimal communication model  introduces only a small amount of additional error (i.e., compare \texttt{z\_comm} and \texttt{z\_true} in Table \ref{tab:flight}).
Furthermore, all the takeaways together indicate our extensive UGV tests are sufficient to fairly evaluate our approach in Section \ref{sec:ugv-experiments}.

\subsection{UGV Experiments}
\label{sec:ugv-experiments}

\bns
\textbf{Hardware:}
Three agents, $A$, $B$ and $C$, were used (see Figure \ref{fig:ugv-system-diagram}). All are Turtlebot2 UGV with 6 UWB sensors evenly distributed in the $xy$-plane \meas{0.32m} from the center. Agent $A$, $B$, and $C$'s sensors are mounted \meas{1.75m}, \meas{0.5m}, and \meas{0.5m} above the ground respectively (i.e., $z\AB = -1.25$m).

\bs
\textbf{Trials:} A full 22 datasets were collected with all three agents, each with 6 antennas, moving in various ways within the mocap space. Ranging measurements are performed by each pair of antennas between agents at \meas{25Hz}. Together the datasets total to nearly \meas{6h} (effectively creating over \meas{200h} of pairwise measurements). These datasets were used for Figures \ref{fig:long-tail} and \ref{fig:noise-el}. Tables \ref{tab:pos-err} and \ref{tab:yaw-err} show results for the 5 trials where all three agents are continuously moving in various arbitrary patterns with max positional and angular velocities of \meas{1m/s} and \meas{1rad/s} respectively. Additionally, two outdoor trials were conducted without mocap  to verify the system generalizes to non-laboratory spaces and can operate for extended periods of time.

\bs
\textbf{Parameters:}
(1)~Equation \ref{eq:opt-f} is solved by \texttt{trust-constr} as provided by \texttt{scipy}'s \texttt{minimize} function \cite{virtanen2020scipy}.\footnote{As in \cite{DBLP:conf/iros/FishbergH22}, initialized with previous estimate if available, otherwise $\tup{\texttt{avg}(\dvm\AB),0,0}$. Here $\texttt{avg}(\dvm\AB)$ is the mean of the current measurements.} State estimation is run at a fixed 1Hz.\footnote{Each robot's onboard i7-4710HQ processor can run estimation at \mytexttilde5Hz serially and >25Hz in parallel (i.e., faster than the UWB sensor rate).}
(2)~Our mean bias correction, $\dn\AiBj(\TAB)$, is the learned 6-degree polynomial with respect to elevation shown in Figure~\ref{fig:noise-el}.
(3)~The current range measurements, $\dm\AiBj$, and pose estimates are smoothed by a \meas{1s} and \meas{4s} moving average filter respectively.
(4)~Our loss function, $\L$, is selected to be the Huber loss $\rho_\delta(a)$ with $\delta = 0.06$.
As with \cite{zheng_uwb-vio_2022,zhang_range-aided_2023,DBLP:journals/corr/abs-2103-04131}, Huber loss was selected due to its reduced outlier sensitivity \cite{hartley2003multiple}.

\bs
\textbf{Discussion of Results:} Results are shown in Tables \ref{tab:pos-err} and \ref{tab:yaw-err}, Figure \ref{fig:ugv-plots}, and Figure \ref{fig:pos-err-vs-alt-err}. To demonstrate the value of individual algorithmic decisions, the tables toggle \texttt{el\_bias}, \texttt{z\_fixed}, \texttt{Huber}, where the final row (red) represents the proposed approach.
Specifically, a check indicates:
\begin{itemize}
\item \texttt{el\_bias}: $\dn\AiBj(\TAB)$ is the learned mean bias correction (see Figure \ref{fig:noise-el}), otherwise $0$.
\item \texttt{z\_fixed}: $z$ is constrained to $\z\AB$, otherwise free.
\item \texttt{Huber}: $\ell(a) \triangleq \rho_{0.06}(a)$, otherwise squared error loss.
\end{itemize}

\noindent Our data shows the proposed approach (red) provides a 9$\times$ improvement over a direct NLLS trilateration (purple), similar to that used by \cite{xianjia_cooperative_2021}. The addition of the \texttt{el\_bias} improves our approach's mean APE by an average of 19\% (i.e., red vs orange). Overall, constraining $z$ leads to the largest APE gains (as predicted by Table \ref{tab:theoretical-stdev}). Unlike APE, there is not a clear best approach for AHE, but the proposed (red) is only at most $1^\circ$ behind the best for any given trial. Additionally, Figure \ref{fig:pos-err-vs-alt-err} justifies our minimal communication protocol, demonstrating that even when operating with \textit{maximal} altitude error within the experimentally demonstrated error tolerances (i.e., $\pm 0.10$m from Figure \ref{fig:altitude}), red outperforms other formulations. Finally, the two outdoor trials, highlighted in the video, demonstrated comparable position accuracy to the laboratory trials as well as the long-term stability and reliability of the system.

\section{Conclusion}
\label{sec:conclusion}

This work presents a UWB-ranging inter-agent relative pose estimation system that minimizes communication load.
Our work outperforms the most-similar recent work \cite{xianjia_cooperative_2021} (APEs between \meas{0.40m} and \meas{1m}). We also remain competitive with other state-of-the-art approaches that have significantly higher communication costs (APE and AHE on the order of \meas{0.10m} and $1^\circ$ respectively), whose heavy reliance on \textit{continuously} transmitted measurements could lead to scaling issues with increased team size and/or decreased communication network capability.

By comparison, this work achieves a mean APE and AHE of \meas{0.24m} and $9.5^\circ$ respectively across our experimental trials, nearly a 9$\times$ improvement over a direct NLLS trilateration approach \textbf{while requiring no \textit{continuously} transmitted measurements}.
Our learned measurement mean bias error correction, $\dn\AiBj(\TAB)$, greatly improves our APE by an average of 19\% across our experimental trials. In the future, this bias correction can be further improved as a neural network, potentially dependent on more than just relative elevation -- that being said, a key takeaway is that \textbf{most UWB systems would benefit from a pose-dependent UWB ranging error model.} Furthermore, future work will include additional flights on multiple larger drones in expansive outdoor environments, integration of this system into a larger resource-aware distributed SLAM pipeline, and locally synchronizing UWB hardware as in \cite{sadowski2023position}.

\bibliographystyle{one-author}
\bibliography{mybib}{}

\clearpage
\appendices

\end{document}